\documentclass[]{article}
\usepackage
[
a4paper,% other options: a3paper, a5paper, etc
left=3cm,
right=3cm,
top=4cm,
bottom=4cm,]
{geometry}
\usepackage[numbers]{natbib}
\usepackage{epsfig,amsfonts}
\usepackage{natbib}
\usepackage{amsmath}
\usepackage{amssymb}
\usepackage{amsthm}
\usepackage{array}
\usepackage{algorithm}
\usepackage[noend]{algpseudocode}
\usepackage{setspace}
\usepackage{multirow}
\usepackage{caption}
\usepackage{subcaption}
\usepackage{pdfpages}

\definecolor{redcolor}{RGB}{255,0,0}
\newcommand{\red}[1]{\textcolor{redcolor}{#1}}

\title{Deformable Classifiers}
\author{
Jiajun Shen\thanks{Jiajun Shen is with the Department of Computer Science at the University of Chicago. Email: jiajun@cs.uchicago.edu}
\ and
Yali Amit\thanks{Yali Amit is with the Departments of Statistics  at the University of Chicago. Email: amit@marx.uchicago.edu.}
}

\begin{document}

\maketitle
\begin{abstract}

Geometric variations of objects, which do not modify the object class, pose a major challenge for object recognition. These variations could be rigid as well as non-rigid transformations. In this paper, we design a framework for training deformable classifiers, where latent transformation variables are introduced, and a transformation of the object image to a reference instantiation is computed in terms of the classifier output, {\it separately for each class.} The classifier outputs for each class, after transformation, are compared to yield the final decision. 
As a by-product of the classification this yields a transformation of the input object to a reference pose, which can be used for downstream tasks such as the computation of object support.
We apply a two-step training mechanism for our framework, which alternates between optimizing over the latent transformation variables and the classifier parameters to minimize the loss function. We show that multilayer perceptrons, also known as deep networks,  are well suited for this approach and achieve state of the art results on the rotated MNIST and the Google Earth dataset, and produce competitive results on MNIST and CIFAR-10 when training on smaller  subsets of training data. 
%\red{(Should we just say deep networks, instead of limiting the architectures to multilayer perceptrons? We use CNNs for our %experiments anyway.)}
\end{abstract}
\section{Introduction}
Ulf Grenander pioneered the idea of handling the challenge of geometric variability of objects in images
in a generative framework using deformable templates \cite{grenander-chow-keenan,amit-jasa,grenander-book,grenmil}.
The variability in a population of images is modeled via deformations applied to a prototype or template. 
The deformations are explicitly parameterized and represented by latent unobserved random variables. 
The statistical framework yields a cost function that measures the distance between the deformed template and the data. This typically  has the form of a sum of squares or other likelihood based measure, usually assuming conditional independence of the pixel observation given the latent variable. There are then two interrelated challenges:  given a template compute the deformation conditional on an image and estimate a template from a sample of images. The first problem has been studied very extensively in a wide variety of contexts, see for example \cite{MTY,younes}. The problem of template estimation has received
some attention, see \cite{amit-trouve, alla}.

In this paper we extend these ideas  
from the domain of generative models to that of discriminative models. 
Given a classifier one can try to compute for each class and image the deformation yielding the optimal output for that class,
and then label the image as the class with the highest output. 
%\red{In other words, instead of the distance between the template of the class
%and an image, we use the output of the classifier on that class. 
In other words, instead of the distance between the template of the class and an image, our framework uses a cost function based on the class scores computed by the classifier.
Furthermore, given samples of images from the different classes one can train
a classifier by iterating the following two steps: first find the optimal deformation of each image to the different classes given the current parameters of the classifier and then update the parameters of the classifier given the optimal deformations of the training images to each class.
Since this is most easily formulated using gradient descent type optimization, we need classifiers that are both deformable with respect to 
their parameters and with resect to their input. Multilayer perceptrons also known as deep networks are then a natural choice.

Deep networks have been very successful in recent years in a wide range of classification and detection tasks. The expressiveness of these networks allows the models to explore possible variations in the data and learn visual representations that are robust to task-irrelevant variations. However, such variations need to be observed in the data when training the networks, without special design the networks do not generalize to unobserved variations in the data. 
The standard solution for improving the transformation-invariance of the classifiers is data augmentation,
where transformed versions of the original data are generated and added to the original data \cite{fasel2006rotation, dieleman2015rotation, laptev2016ti, dieleman2016exploiting, van2017learning}. This approach works well but we believe it
is of interest to explore the alternative where an explicit computation of the latent deformations is performed during the classification.
In theory this means that the classifier does not need to be as flexible since it is not directly trying to discriminate between many different
instantiations of the classes, rather it only needs to learn to discriminate between images of the objects at reference instantiation. Furthermore, obtaining the reference pose of the object as part of the output of classification can assist in additional visual tasks.
One such task is determining the support of the object in the image.

Spatial transformer networks (STN) \cite{jaderberg2015spatial} also transform the image as part of the classification process.
They try to remove extraneous transformation variability a priori by introducing a spatial transformation module before the classification network. The image data is {\it first} transformed to a reference instantiation, independent of the class, and then passed on to be classified. The information the network uses to first transform the image is independent
of the class and therefore is necessarily generic. We believe it is essentially computed based on the first and second order statistics of the pixel data, and as we show it is very sensitive to clutter. It is, however, of interest that the transformation network and the subsequent classification network are trained together end to end using gradient descent. 

Other approaches to transformation invariance explicitly transform the network filters so that the feature maps are invariant to the selected types of transformations \cite{wu2015flip, teney2016learning, marcos2016rotation, zhou2017oriented}. Another family of approaches tries to generalize convolutional architectures by either extending the feature space to a group space of transformations \cite{gens2014deep, cohen2016group}, or warping the input so that the transformation equivariance is implicitly encoded \cite{henriques2016warped}. These approaches are either limited to a small set of transformations, or they need to keep the models shallow because of the high computational burden required to consider additional transformations in the feature map.

%We ask whether it is possible to use the classifier itself to remove extraneous transformation variability from the data and obtain reference poses of the objects.

In our framework, latent variables are introduced, separately for each class, to capture the transformations of the data and we apply a two-step training mechanism to alternatively optimize over the latent variables and the neural network model parameters to minimize a designed loss function. We emphasize that the latent variables are optimized for each class separately. Consequently, unlike STN that produces a single transformed version of the original input, here we produce a transformed version for each class. The transformation is not predicted directly from the data, rather, for each class it is estimated to optimize the output of the unit representing that class.

We show that this framework can be applied to any existing neural network architecture and offers flexibility in the types of transformation considered by the model. We apply our framework to the training of convolutional neural networks (CNN), and present competitive results on mnist, mnist-rot, CIFAR-10, and the Google Earth dataset. In addition to improved classification rates, we show that the
estimated latent transformations indeed align the images very well, and allow us to estimate very precise object supports in the case of
mnist, and the correct object rotation in the case of Google Earth.

In Section \ref{related} we describe related work on latent variables in the machine learning literature. In Section \ref{DC} we layout the deformable classifier algorithm. In Section \ref{section_CSTN} we describe a modification of the spatial transformer network
that regresses the parameters of the transformation on the image as the STN but separately for each class. In Section \ref{exp} we describe the experiments and in Section \ref{clutter} we show how these types of networks can be used to handle clutter in the case of
handwritten digits.

\section{Related work}\label{related}
\begin{figure}[]
	\centering
	\includegraphics[width=0.60\textwidth]{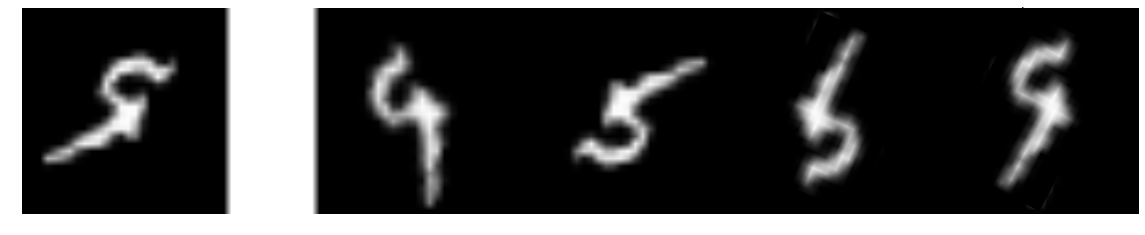}
	\caption{A rotated image of digit five can be further rotated to look like instantiations from digit class four, five, six and nine.}
	\label{spn_bad_example}
\end{figure}

Two approaches most relevant to our proposed method are spatial transformer networks and latent SVM models.
\def\TT{\mathcal{T}}
\def\xx{\textbf{x}}
\def\zz{\textbf{z}}

\subsection{Spatial Transformer Network (STN)} The model of \cite{jaderberg2015spatial} consists of a spatial transformer module that contains a localization network and a grid generator together with a classification network that can be trained end to end using stochastic gradient descent. The localization network predicts the transformation parameters based on the input image. These could be the six parameters of an affine map or a more general smooth deformation described for example through a thin plate spline. 
The grid generator transforms the image and the resulting transformed image is passed through the classifier.
The transformation of the image is defined in a `weak' sense as follows.
Consider the image domain $D$ as a continuum where the image is defined  as $\xx(s), s\in D$. Given a parameterized family of smooth deformation functions $\phi(s,\zz)$, mapping $D$ to $D$ the deformed image associated with $\phi$ can be expressed as $\TT_\zz \xx(s) = \xx(\phi(s,\zz))$.
This is the original formulation in \cite{amit-jasa}.  Now take a smooth kernel function $K(t,u)$ defined on $D \times D$, which approximates the Dirac delta function and write
\begin{align}
	\label{prototype_deformation}
	\TT_\zz \xx(s) \sim \int_{u\in D} \xx(u)K(\phi(s,\zz), u)\,du
\end{align}
The derivative with respect to $\mathbf{z}$ becomes:
\begin{align}
\label{derivative_deformation}
	\frac{\partial \TT_\zz \xx(s)}{\partial \mathbf{z}} =  \int_{u\in D} \xx(u)\frac{\partial K}{\partial t}(\phi(s,\mathbf{z}), u)\,du\frac{\partial\phi(s,\mathbf{z})}{\partial\mathbf{z}}.
\end{align}
This formulation allows to push the application of the deformation and the computation of derivatives onto the smooth kernel (which can be done analytically) and avoids the need to deal with explicit deformations or derivatives of the image, which is defined on a discrete pixel grid. Denote the network
predicting the transformation parameters $\zz$ as $\Psi(\xx,\eta)$ and the subsequent classification neural network $\Phi(\cdot,\theta)$. The entire system is defined as
 $\Phi(\TT_{\Psi(\xx,\eta)} \xx (s),\theta).$ The gradients with respect to $\theta$ and $\eta$ are easily propagated backwards through
 this network provided a module is defined to compute $\phi(s,\zz)$.
Once the network is trained, i.e $\hat \eta$ and $\hat\theta$ are estimated, an image $F$ is passed through $\Psi$ to obtain the predicted transformation $\zz=\Psi(\xx,\hat \eta)$. Then $\phi(s,\zz)$ is computed yielding $\TT_\zz \xx(s)$  and $\Phi(\TT_\zz \xx(s),\hat \theta)$ gives the classification.

In STN \cite{jaderberg2015spatial}, the authors use the bilinear sampling kernel 
$$K(t,u)=\max(0,1-|t_0-u_0|)\max(0,1-|t_1-u_1),$$
and Equation(\ref{prototype_deformation}) is approximated as follows:
\begin{align}
	\label{spn_prototype_deformation}
	\TT_\zz \xx(s) = &\sum_{u\in D}\xx(u)\max(0, 1-|\phi(s,\zz)_0 - u_0|)\max(0, 1-|\phi(s,\zz)_1 - u_1|) \nonumber \\  = &
	\sum_{| u - s| < 1} |\phi(s,\zz)_0 - u_0| \cdot |\phi(s,\zz)_1 - u_1|
	, s \in D.
\end{align}

Note that the spatial transformation is computed directly from the image without knowing its class. This can create
ambiguities such as in Figure \ref{spn_bad_example} where we show an image of rotated digit five and how it can be rotated to look like instantiations of different digit classes. Intuitively, the spatial transformer module should recognize the class label of the image, and then extract the transformation parameters taking the class label into account. We argue that an accurate estimate of transformation is not possible unless the image label is captured, and even given the class, one expects that some external information is needed to guide the transformation. In STN the class labels of the images are not fully captured by the spatial transformer module $\Psi$ (otherwise we would not need a downstream network to handle the classification task) and the spatial transformer module cannot produce accurate spatial transformation for the input.

\subsection{Discriminative Latent Variable Models:} Latent variables in discriminative models have been studied in the framework of multiple instance learning (MIL), where the latent variables are used to capture the variations of instances within the same labeled bag. The MI-SVM formulation of multiple instance learning was initially proposed in \cite{andrews2003support}, and later reformulated as latent SVM in \cite{felzenszwalb2010object}. In that work, the problem is to detect objects of a given class, and a binary classifier is run across the image scoring each window $x$ as follows:
\begin{equation}
	f_{\beta}(\mathbf{x}) = \max_{\mathbf{z}}\beta \cdot \Phi(\mathbf{x, z})
	\label{binary_score_function}
\end{equation}
Here $\beta$ is a vector of model parameters, $\Phi(\mathbf{x})$ is the feature extraction function for $\mathbf{x}$ and $\mathbf{z}$ are latent values.  %Hidden CRF \cite{quattoni2007hidden} is also closely related, where it marginalizes over the latent variables instead of maximize over them.}
In this model the only classifier parameters are $\beta$, and a hinge loss is used to train the model:
\begin{equation}
L(\beta) = \frac{1}{2} \Vert \beta \Vert^2 + C \sum_{i=1}^n \max(0,1-y_i f_\beta(\mathbf{x})),
\end{equation}

where $y_i=1,-1$ for object and non-object examples respectively.
Since $f_{\beta}$ is a maximum of linear functions it is convex in $\beta$ and so for negative examples
for which $y_i=-1$ the summand in the cost function is convex in $\beta$. For positive examples it is not convex 
and the authors propose a two step iteration. For each positive example find the optimal $z^*_i$ given the current
value of $\beta$ and then optimize the convex function

\begin{equation}
\tilde L(\beta) \\=  \frac{1}{2} \Vert \beta \Vert^2 + C \sum_{y_i=1} \max(0,1-\beta \cdot \Phi(\mathbf{x_i, z^*_i})) + C \sum_{y_i=-1}
\max(0,1+ \max_{\mathbf{z}}\beta \cdot \Phi(\mathbf{x_i,z})).
\end{equation}

In a similar vein, in this paper we incorporate latent variables into deep neural networks to capture the transformations of the input. The input image is warped for each class separately based on the latent values that optimize its output on that class. In training, this is done for the entire batch using the old network parameter values, and then one or more gradient steps are taken to update the network parameters after the input images are warped. In our setting the classifier is not linear in its parameters and we will have no choice
but to compute the optimal latent variable for each class for all examples.

\section{Deformable Classifiers}\label{DC}
\def\jj{\mathbj{j}}
We consider a multi-class classifier on $C$ classes that scores an example $\mathbf{x}$ for each class as follows:	\begin{equation}
	f_{\beta_\mathbf{j}}(\mathbf{x}) = \max_{\mathbf{z}}\beta_\mathbf{j} \cdot \Phi_\theta(\mathcal{T}_{\mathbf{z}}\mathbf{(x)}), j=1,\ldots,C.
	\label{score_equation}
\end{equation}
Here $\beta_\mathbf{j}$ is a vector of model parameters for class $\mathbf{j}$, $\Phi_\theta$ is a feature mapping function parametrized by $\theta$, which in our setting is a multi-layer convolutional neural network. The paramters $\theta$ are common to all classes
and the final classification depends on the last layer of the network that is parameterized with the $\beta$'s. The latent variable $\mathbf{z}$ is introduced to parametrize the deformations of the data and $\mathcal{T}_{\mathbf{z}}(\cdot)$ transforms the input according to the value of $\mathbf{z}$. The label $\hat{c}$ of each example is then determined by
\begin{equation}
	\hat{c} = \arg\max_\mathbf{j} f_{\beta_\mathbf{j}}(\mathbf{x}).
\end{equation}
For a test example, the model finds a separate optimal latent value for each class in terms of the output corresponding to that class. The class output with highest value yields the final classification. Together with that classification, we also obtain an optimal transformation of the image into reference pose.

Intuitively, in order to make the correct prediction, we want the score of the target class to be larger than the scores of the non-target classes. As a result, we can optimize over the model parameters and maximize the margin between $f_{\beta_\mathbf{y}}(\mathbf{x})$ and $f_{\beta_\mathbf{j}}(\mathbf{x})$ for $\mathbf{j} \neq \mathbf{y}$. Suppose we have a set of observations $\mathbf{x} = \{\mathbf{x_1, \dots,x_N}\}$ and the corresponding data labels $\mathbf{y} = \{\mathbf{y_1, \dots,y_N}\}$, we use the multiclass hinge loss as follows: 
\begin{equation}
	\mathcal{L}(\Theta) = \sum_\mathbf{i} \max(0, 1 + \max_{\mathbf{j} \neq \mathbf{y_i}}f_{\beta_\mathbf{j}}(\mathbf{x_i})- f_{\beta_{\mathbf{y_i}}}(\mathbf{x_i})) + \lambda(\lVert\theta\rVert^2 + \sum_\mathbf{j}\Vert\beta_\mathbf{j}\rVert^2)
	\label{loss_equation}
\end{equation}
where $\Theta = \{\theta, \beta_1, \dots, \beta_C\}$ are the model parameters and $C$ is the number of classes. $\lambda$ is the parameter that controls the regularization term $\lVert\theta\rVert^2 + \sum_\mathbf{j}\Vert\beta_\mathbf{j}\rVert^2$.

The key step in our method is to find the optimal instantiation of each example for each class. At first glance, it might seem that it would be simpler to forgo the non-target classes and only focus on finding the optimal instantiation of the example for the target class. We note, however, that such an approach is often insufficient. Recall that an image can be transformed to look like instantiations from a non-target class, like the examples we show in Figure \ref{spn_bad_example}. Without competing with optimal instantiations of the data from non-target classes, the model might not be learning from the most competitive negative examples. 
%Moreover, as opposed to finding one optimal instantiation of the example for all classes, optimizing for each class separately makes the problem easier and leads to more accurate estimates of the optimal instantiations.

In order to minimize the hinge loss in Equation (\ref{loss_equation}), we design a two-step training mechanism. For each example, the algorithm finds the highest scoring latent values for each class based on the current model parameters. Then the algorithm optimizes over the model parameters while fixing the latent values.
%, and it can maximize the margin between $f_{\beta_\mathbf{y_i}}(\mathbf{x_i})$ and $\max_{\mathbf{j} \neq \mathbf{y_i}}f_{\beta_\mathbf{j}}(\mathbf{x_i})$. 
%When optimizing over the model parameters, an iterative approach such as gradient descent can be applied to update the parameters for multiple iterations. 
We outline the procedure for the two-step training algorithm in Algorithm \ref{EM-algorithm} for deformable classifiers (DC).
\begin{algorithm}[h]
	\caption{Two-Step Algorithm For Learning a Deformable Classifier (DC)}\label{EM-algorithm}
	\begin{spacing}{1.5}
		\fontsize{10}{10}\selectfont
		\begin{algorithmic}[1]
			\Procedure{}{}\\
			\ \ \ \ \ \ Choose an initial seting for the parameters $ \Theta^{\text{old}} = \{ \theta^{\text{old}}, \beta_1^{\text{old}}, \dots, \beta_C^{\text{old}}\}$.\\
			\ \ \ \ \ \ \textbf{Optimize Over $\mathbf{z}$:}\\
			\ \ \ \ \ \ \ \ \ \ \ \ $f_{\beta_\mathbf{j}}(\mathbf{x_i})= \max_\mathbf{z} \beta_{\mathbf{j}}^{\text{old}}\cdot\Phi_{\theta^{\text{old}}}\left(\mathcal{T}_\mathbf{z}\left(\mathbf{x_i}\right)\right), \ \mathbf{j}=1,\ldots,C$.\\
			\ \ \ \ \ \ \ \ \ \ \ \ $\mathcal{L}(\Theta) = \sum_\mathbf{i} \max(0, 1 + \max_{\mathbf{j} \neq \mathbf{y_i}}f_{\beta_\mathbf{j}}(\mathbf{x_i})- f_{\beta_{\mathbf{y_i}}}(\mathbf{x_i})) + \lambda(\lVert\theta\rVert^2 + \sum_\mathbf{j}\Vert\beta_\mathbf{j}\rVert^2)$\\
			
			\ \ \ \ \ \ \textbf{Optimize Over Model Parameters $\Theta$:}\\
			\ \ \ \ \ \ \ \ \ \ \ \ $\theta, \beta_1, \dots, \beta_C = \arg\min_{\theta, \beta_1, \dots, \beta_C}\mathcal{L}(\Theta)$\\
			\ \ \ \ \ \ \textbf{If} \ the convergence criterion is not satisfied \ \textbf{then}\\
			\ \ \ \ \ \ \ \ \ \ \ \ $\theta^{\text{old}}\leftarrow\theta^{\text{new}}$, $\beta_1^{\text{old}}\leftarrow\beta_1^{\text{new}}$, \dots, $\beta_C^{\text{old}}\leftarrow\beta_C^{\text{new}}$\\
			\ \ \ \ \ \ \ \ \ \ \ \  and return to line 3.\\
			\ \ \ \ \ \ \textbf{Stop}
			\EndProcedure
		\end{algorithmic}
	\end{spacing}
\end{algorithm}

When the latent variables form a discrete set, for example a finite set of rotations, optimization is performed by exhaustively search (DC-ES). For continuous latent variables, we optimize $f_{\beta_\mathbf{j}}(\mathbf{x, z})$ from Equation (\ref{score_equation}) with respect to $\mathbf{z}$ by gradient descent (DC-GD). We regularize the magnitude of $\mathbf{z}$ during optimization by penalizing its distance from the identity. 
When the range of the continuous variable is very large, such as the 360 degree range
of rotations, we initialize the gradient descent at a small set of discrete initial rotations and take that optimal value over all initializations (DC-ESGD).

In this paper, we use 2D affine transformations and thin plate spline transformations \cite{bookstein1989principal} as the two parameterizations for $\phi$ as the gradient of $\phi(s,\zz)$ with respect to $\zz$ for these two types of transformations has been implemented efficiently in the deep learning package Lasagne which we employ for our experiments.

\section{Class Based Spatial Transformer Network}\label{section_CSTN}

\begin{figure}[h]
	\centering
	\includegraphics[width=0.8\textwidth]{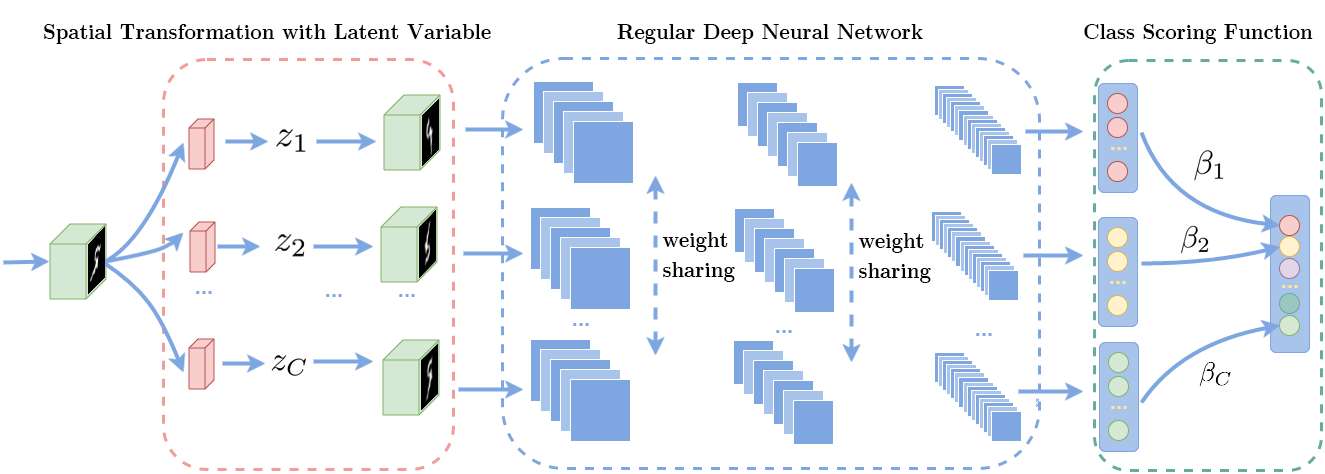}
	\caption{Model architecture for CSTN.}
	\label{latent_svm_model}
\end{figure}
%% Need to clarify some things here.
A different approach, which is a direct generalization of the original spatial transformer model, is to use class based spatial transformer modules (CSTN), one for each class, to directly predict values of $\mathbf{z}$ for each class during training and testing. Instead of optimizing over $\mathbf{z}$ based on the gradient of the classifier output, during training this approach optimizes over the neural network parameters that predict the class-specific transformation. Each transformation of the image is then passed through the same feature extraction network, and the classifier calculates the class scores based on the features extracted from different class-specific transformations of the image, see Figure \ref{latent_svm_model}. To be precise let $\Psi_c(\xx,\theta_c)$ denote the network computing the transformation for each
 class $c$. Let $\Phi(\cdot,\eta)$ be the common feature extraction network then the full CSTN computes
\begin{equation}\label{CSTN}
\arg\max_c \beta_c \cdot \Phi(\TT_{\Psi_c(\xx,\theta_c)}\xx,\eta).
\end{equation}

%In training, as in the previous section, we alternate between updating the parameters $\eta, \beta_c, c=1,\ldots,C$ of the classification
%network with 
%the $\theta$ parameters fixed, and then keeping the classification network parameters fixed and updating the $\theta$ parameters of the transformation networks $\theta_c,c=1,\ldots,C$.
%When $\theta$ is fixed the loss for example $\xx_i$ is given by 
%$$L(\xx_i,y_i,\eta,\beta) = S\left(y_i, \left[\beta_c \cdot \Phi(\TT_{\Psi_c(\xx_i,\theta_c)}\xx_i,\eta)\right]_{c=1}^C \right),$$
%where $S$ could be the softmax loss or the hinge loss described above. When $\eta$ and $\beta$ are fixed
%the loss for example $\xx_i$ is based only on the output for class $y_i$
%\red{
%$$ L(\xx_i,y_i,\theta_{y_i}) = \beta_{y_i} \cdot \Phi(\TT_{\Psi_{y_i}(\xx_i,\theta_{y_i})}\xx_i,\eta),$$
%(Is $y_i$ the ground-truth label of $x_i$? When $\eta$ and $\beta$ are fixed, we update the parameters not only for $\theta_{y_i}$, but also for other non-target classes $\theta_c$, $c = 1, \dots, C$. By doing this, we are trying to find the most competitive negative examples. I am rewriting this paragraph below.)}
%so the only parameter influenced by this example is $\theta_{y_i}$.
%In this state the network tries
%to modify $\theta_{y_i}$ to make the image $\xx_i$ look
%as close as possible to a standard image of an object of class $y_i$. 

In training, as in the previous section, we alternate between updating the parameters $\eta, \beta_c, c=1,\ldots,C$ of the classification
network with 
the $\theta$ parameters fixed, and then keeping the classification network parameters fixed and updating the $\theta$ parameters of the transformation networks $\theta_c,c=1,\ldots,C$.
When $\theta$ is fixed the loss for example $\xx_i$ is given by 
$$L(\xx_i,y_i,\eta,\beta) = S\left(y_i, \left[\beta_c \cdot \Phi(\TT_{\Psi_c(\xx_i,\theta_c)}\xx_i,\eta)\right]_{c=1}^C \right),$$
where $S$ could be the softmax loss or the hinge loss described above. When $\eta$ and $\beta$ are fixed, we want to optimize the spatial transformer modules $\Psi_c(\xx,\theta_c)$ separately for each class $c$, i.e: we want to maximize the following
	$$ L(\xx_i,c,\theta_{c}) = \beta_{c} \cdot \Phi(\TT_{\Psi_{c}(\xx_i,\theta_{c})}\xx_i,\eta),$$
so for each transformation network $\Psi_c(\xx, \theta_c)$, we optimize the parameter $\theta_c$ to compute the optimal instantiation of each example for that class and make the $\xx_i$ look as close as possible to an image of an object of class $c$. As a result, the spatial transformer modules not only learn how to find the optimal instantiation of each example for the target class, but also learn to compute competitive negative examples for the non-target classes.

In testing there is no need for optimization as the transformation for each class is predicted directly through $\Psi_c$. Again,
unlike STN where only one spatial transformer module is trained for the network, this approach constructs a different spatial transformer module for each class, providing a different latent value for each class. The downstream networks from each transformer module are all tied until the final layer, where each one feeds into the corresponding class output unit. 
The methods DC-GD, DC-ESGD and CSTN require us to parametrize the transformation function in a form that is differentiable with respect to the latent variable so that we can use the gradient to either directly update the latent variable or update the model parameters in the spatial transformer modules.

%Another parameterization we use for the paper is thin plate spline transformation\cite{bookstein1989principal}, where a set of control points are introduced and arranged as a grid for the image. A much complicated transformation is constructed by allowing different offsets for the control points as well as imposing minimization of an energy function to enforce smoothness and rigidness of the deformation. 

%These latent values are then used to transformed the input according to the deformation parametrization.
%Depending on which approach we use to obtain $\mathbf{z}$, the module for spatial transformation marked as the red region in Figure \ref{latent_svm_model} involves a different structure and procedure. For CSTN the input will be passed to separate localization networks  to predict latent values. For ES, GD and ESGD, a different procedure is used: the model transforms the input based on some initial values of $\mathbf{z}$, then passes the transformed input to the rest of the network to calculate the class scores or gradients so that the the latent values of $\mathbf{z}$ can be adjusted accordingly. This procedure will be repeated multiple times before the module finds the optimal latent values.
\section{Experiments}\label{exp}
We implement our model and perform experiments on the mnist-rot dataset, the CIFAR-10 dataset and the rotation angle estimation task for the Google Earth dataset.
%to show the effectiveness of the model.
\subsection{The mnist-rot Dataset}

The mnist-rot dataset is a variant of the MNIST dataset \cite{larochelle2007empirical} that consists of images from the original MNIST rotated by a random angle from $0^\circ$ to $360^\circ$. The dataset contains 12000 training images and 50000 testing images.

We set the angle of rotation as the latent variable. We choose a CNN architecture which can be trained to achieve a competitive result on the MNIST dataset. The CNN architecture consists of two consecutive convolutional blocks, where each block is composed of a convolutional layer with 32 filters of size $5\times5$ and a maxpooling layer of size $2\times2$. The output of the convolutional blocks is passed to a fully connected layer with 256 units before being fed to the final layer with 10 units. We initialize the weights of the CNN model by training it on a subset of the original MNIST dataset (first one hundred training images of each class). Then we train the DC model with optimal instantiations on the mnist-rot training data. We experiment with three different approaches to optimizing over the latent variable including, DC-ES and DC-ESGD. In DC-ESGD $\mathbf{z}$ is initialized at eight different rotations, and each is optimized for ten iterations using gradient descent. We choose the value of $\mathbf{z}$ that produces the highest score.
\begin{figure}[t]
	\centering
	\includegraphics[width=0.60\textwidth]{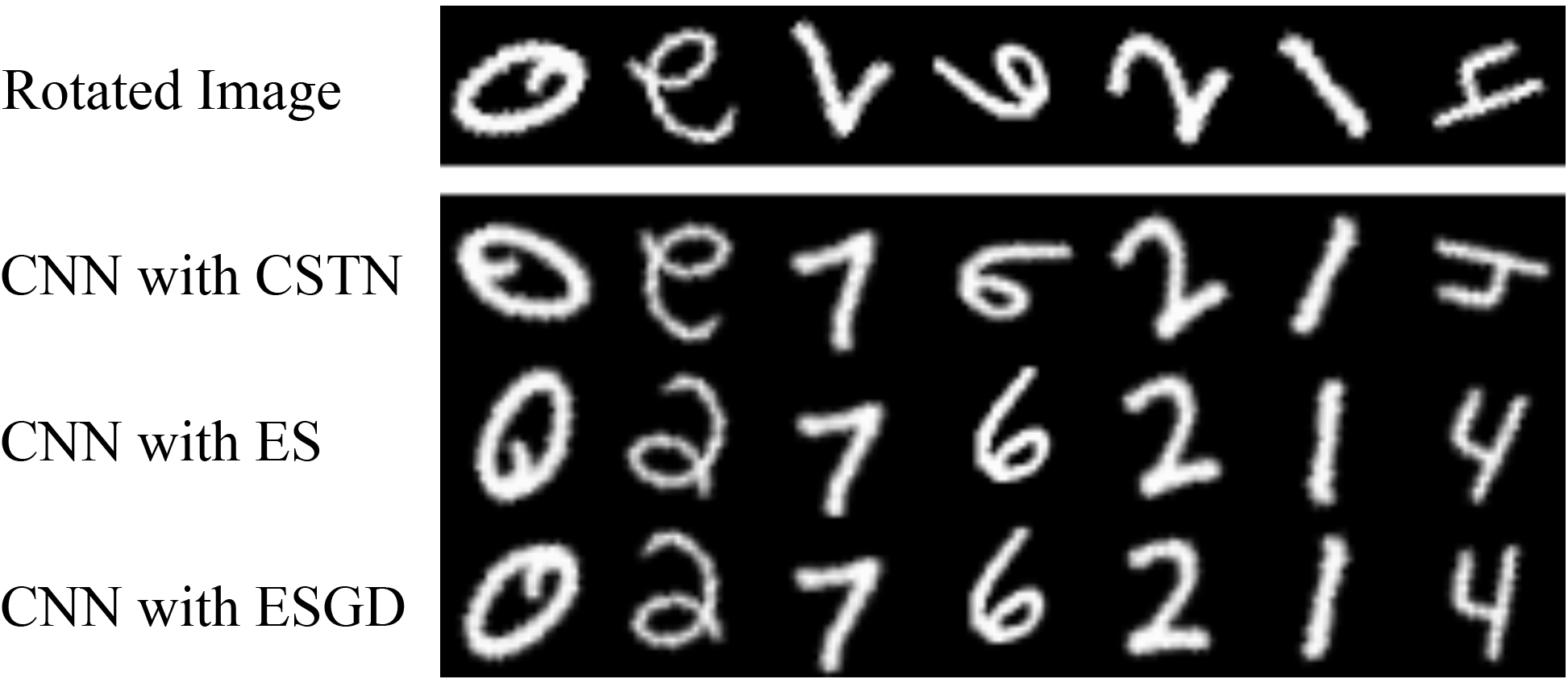}
	\caption{Correct the image rotations using the latent rotation angles estimated by three optimization approaches.}
	\label{gradient_vs_exhaustive_left}
\end{figure}

In Figure~\ref{gradient_vs_exhaustive_left}, we show some example images of rotated digits and their unrotated versions corrected using the latent rotation angles estimated by the three approaches. Compared to CSTN, the DC-ES and DC-ESGD achieve better estimates of the rotated angles. The exhaustive search approach is more constrained since it can only search for a limited amount of rotations (in this case every 45 degrees). The gradient descent approach can adjust the rotation with an arbitrary angle, creating better rotation-corrected images. In Table \ref{gradient_vs_exhaustive_right}, we show the error rate achieved by different models. When using class-specific spatial transformer modules to optimize over the latent variables (CSTN), we are able to achieve an error rate of $2.64\%$, significantly improved from $5.71\%$ achieved by the conventional STN. Our best result is achieved with DC-ESGD, reaching an error rate of $1.25\%$. The state-of-art result $1.2\%$ is achieved by TI-Pooling \cite{laptev2016ti}, where 24 explicitly rotated versions of the images are presented to the model for training and testing. 
\begin{table}[h]
	\centering
	\begin{tabular}{|c|c|} %{m{4.4cm}|m{0.6cm}}
		\hline
		{Model} & Error (\%)\\
		\hline
		TIRBM \cite{sohn2012learning} & 4.2\\
		original CNN Model & 4.1 \\
		STN & 5.71 \\
		TI-POOLING (24 rotations) \cite{laptev2016ti} & \textbf{1.2} \\ \hline
		CSTN & {2.64} \\
		DC-ES (8 rotations)& {2.31}\\
		DC-ESGD (8 initial rotations)&{\textbf{1.25}}\\
		\hline
	\end{tabular}
	\caption{Results on the mnist-rot dataset.}
	\label{gradient_vs_exhaustive_right}
\end{table}

\begin{figure}[b]
	\centering
	\includegraphics[width=0.95\linewidth]{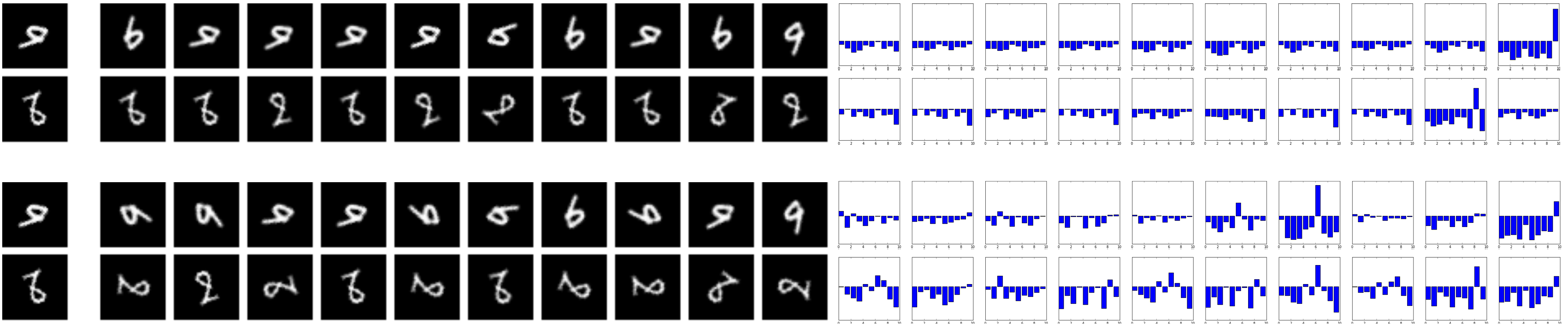}
	\caption{Examples of rotation-corrected images for ten separate classes using a conventional CNN trained on upright digits (bottom) and a CNN trained on rotated digits using DC-ESGD (top). For each rotation-corrected image we show all 10 class scores on the right,
but the classifier only uses the score from the class that determined the rotation.}
	\label{fig:activationvaluecomparison}
\end{figure}

Our training framework allows the model to compare optimal instantiations of the image under different classes and expand the margin between the score of the target class and the highest score of non-target classes. To show why this is important, we conduct the following experiment: We first train a traditional CNN model on 60000 training images of upright digits from the original MNIST dataset with multi-class hinge loss. Then the trained model can be plugged into our framework, and without additional training, we can use it to find the latent rotation angles of the rotated digits under each class. We compare this approach with the DC model trained with optimal instantiations. In Figure \ref{fig:activationvaluecomparison}, we show all ten class scores for each transformed image. Note that the classifier only uses class score $o_c$ from image transformed based on the output for class $c$. 
We see that although the optimal latent rotation angles under the correct class labels captured by the two approaches are similar, the CNN trained using our framework effectively suppresses non-target class scores. While in the examples generated by the conventional CNN, we observe many undesired spikes of  scores for non-target classes, which will lead to incorrect classifications. For DC-ESGD the example rotated 9 has a strong output for class 9 in the last display and very low outputs for any of the other class it has been transformed to. Using the standard CNN the output for class 6 when rotated for class 6 is higher than the output for class 9 when rotated for class 9.
A similar problem occurs with the rotated 8.
Using the standard CNN to rotate the images achieves an error rate of $11.04\%$, which is far worse than any result we shown in Table \ref{gradient_vs_exhaustive_right}.

\subsection{MNIST}

\begin{table}[h]
	\centering
	\begin{tabular} {|c|c|}
		\hline
		{Model}	 & {Error (\%)} \\ \hline
		CNN & 3.17 \\ \hline
		%CNN Model (data augmentation) & 2.48 \\ \hline
		STN & 4.9 \\ \hline
		DC-GD (Thin Plate Spline) & \textbf{2.00} \\
		\hline
	\end{tabular}
	\caption{ Results on MNIST-100.}
	\label{mnist_tps_left}
\end{table}

\begin{figure}[h]
	\centering
	\includegraphics[width=0.6\textwidth]{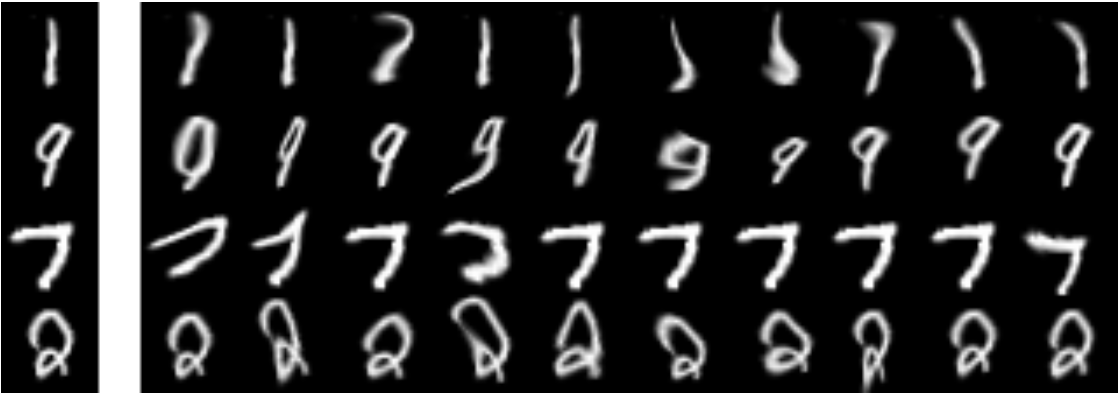}
	\caption{Examples of optimal images deformed by the thin plate spline transformation for different classes. The original images are shown in the first column.}
	\label{mnist_tps_right}
\end{figure}

We  train our model on the original MNIST dataset \cite{lecun1998mnist}. In order to limit the transformation invariance that can be learned from the data, we only use the first 100 images of each class from the training dataset (\textit{called MNIST1000}). In order to capture the local deformations of the data, we use the thin plate spline transformation as the latent variables. 
%As the primary source of transformations come from shearing, skewing and local deformation, we use the thin plate spline %transformation to parametrize the coordinate transformations. 
A  4x4 grid of control-points is used for the thin plate spline transformation, resulting in 32 parameters modeling the image deformations.
As before, we first initialize the CNN model by training it on MNIST1000 dataset and then train the model using our framework. In this experiment, we only use gradient descent (GD) to optimize over the latent variables.

%For example, we observe digit one transforms to a digit seven in the first row, and digit nine, digit seven and digit two transform to digit zero, digit one and digit seven respectively. 
In Figure~\ref{mnist_tps_right}, we show the optimal  transformed image by thin plate splines  for each of the ten classes.  As shown in Table~\ref{mnist_tps_left}, we are able to achieve an error rate of 2.0\% using our framework, which is a major improvement compared to results with the original CNN or with STN.

\subsection{CIFAR-10}
\begin{table}[h!]
	\centering
	\label{result_table_cifar4k}
	\caption{Experiment Result on CIFAR-10(400)}
	\renewcommand{\arraystretch}{1.15}
	\scalebox{0.85}{
		\begin{tabular}{|m{8cm}|m{4cm}|m{1.6cm}|}
			\hline
			{Model Description}	 &{Architecture} & {Error (\%)} \\ \hline
			DCGAN (semi-supervised approach) \cite{radford2015unsupervised} & & $26.2$($\pm0.4$)\\ 
			Exemplar-CNN (semi-supervised approach) \cite{dosovitskiy2014discriminative} & 64c5-128c5-256c5-512f & $24.6$($\pm0.2$)\\
			Exemplar-CNN (semi-supervised approach) \cite{dosovitskiy2014discriminative} & 92c5-256c5-512c5-1024f & $23.4$($\pm0.2$)\\
			Steerable-CNN \cite{cohen2016steerable} & 14 layers, 4.4M params & $24.56$\\ \hline
			\multirow{2}{*}{CNN Baseline} &64c3-64c3-128c3-128c3-&\multirow{2}{*}{$28.43$}\\
			&256c3-256f, 1.6M params&\\\hline
			\multirow{2}{*}{CNN with DC-ESGD (Rotation)} &64c3-64c3-128c3-128c3-&\multirow{2}{*}{$27.9$}\\
			&256c3-256f, 1.6M params&\\\hline
			\multirow{2}{*}{CNN with DC-GD (Translation, Scale)} &64c3-64c3-128c3-128c3-&\multirow{2}{*}{$25.53$}\\
			&256c3-256f, 1.6M params& \\ \hline
		\end{tabular}
	}
\end{table}

\red{
\begin{figure}[]
	\centering
	\includegraphics[width=0.85\textwidth]{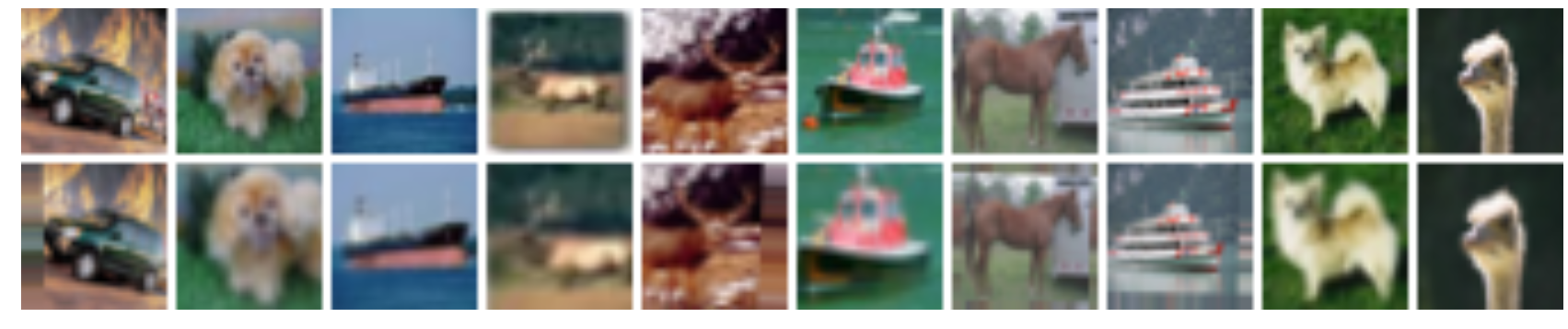}
	\caption{\textit{Top:} Example images from the CIFAR-10 dataset; \textit{Bottom:} Images translated and scaled by our model.}
	\label{cifar10_deformation}
\end{figure}
}

We apply our model on the CIFAR-10 dataset \cite{krizhevsky2009learning}. We train our model using the first 400 images of each class from the training dataset (called \textit{CIFAR-10(400)}) and test on the original CIFAR-10 test dataset. A five-layer CNN model can achieve 28.43\% test error after 4000 epochs of training. We choose a CNN with the same architecture for our framework. We initialize the network by first training the CNN model on CIFAR-10(400)
% REALLY 2000 epochs? 
% Yes, I did train the original CNN for 2000 epochs and then train it using our framework for 2000 epochs. But I think a better way to say this is "train until convergence"
and then train it with the deformable classifier. We explore two settings: one with the angle of rotation as the latent variable for the model and the other with translation and scale as the latent variables. In Figure~\ref{cifar10_deformation}, we show the optimal transformation via translation and scaling for objects  CIFAR-10 images. We show a $2.9\%$ increase of model performance using a CNN with latent translation and scaling, which is even comparable with some of the semi-supervised approaches reported in the literature that use a large complementary unlabeled training set.

\subsection{Google Earth Dataset}
 %We show some examples of training images in Figure \ref{google_earth}.

\begin{figure}[H]
	\centering
	\begin{subfigure}[t]{0.32\textwidth}
		\centering
		\includegraphics[height=1.5in]{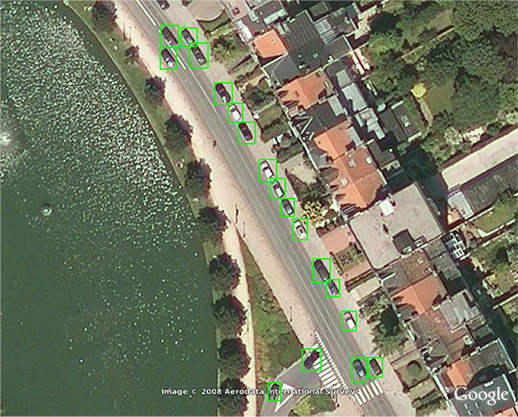}
		\caption{}
	\end{subfigure}%
	~
	\begin{subfigure}[t]{0.32\textwidth}
		\centering
		\includegraphics[height=1.5in]{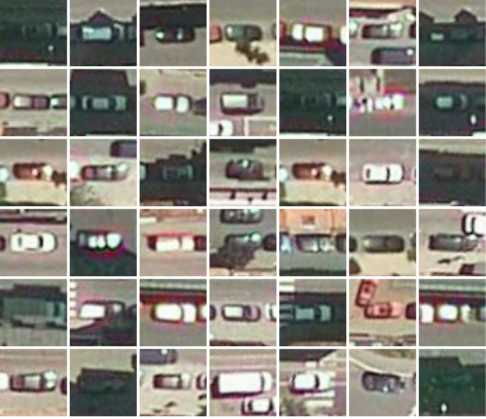}
		\caption{}
	\end{subfigure}
	\begin{subfigure}[t]{0.32\textwidth}
		\centering
		\includegraphics[height=1.5in]{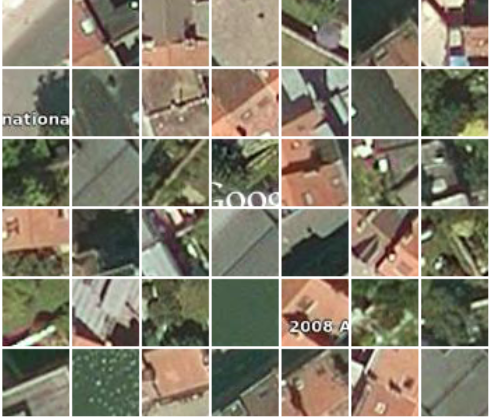}
		\caption{}
	\end{subfigure}
	\caption{(a): An example of training images from the Google Earth dataset. (b) and (c) are examples of car images (car front point to the right) and background images we use for training a detection model for horizontal cars.}
	\label{training_car_image}
\end{figure}

We train our model on the Google Earth dataset \cite{heitz2008learning}, which contains aerial photos of streets with bounding boxes around the vehicles. Henriques et al. \cite{henriques2014fast} also add angle annotation for each vehicle as a supplement of the dataset. The dataset contains 697 vehicles in 15 large images, where the first ten images are used for training and the rest for testing. The task of this dataset is to estimate the rotation parameter for each vehicle in the image.
% Every car in the image is centered, with the car front pointing to the right. 
% We initialize the model with the original CNN model for horizontal car detection, so that the CNN model with latent SVM can capture the preferred orientation of the cars and correct orientations for rotated cars.
We first learn a horizontal car detection model by training a classical CNN model to discriminate between horizontal car images and background images. In Figure~\ref{training_car_image}, we show some image examples for training the detection model. We use this model as initialization to the DC-ESGD training method, which further trains the model using images of rotated vehicles cropped from the training images. Then we can use the trained latent variable model to estimate the rotation angles of the vehicles by finding the latent rotation parameters that give the maximal values for the score function.

\begin{figure}[t]
	\centering
	\includegraphics[width=0.6\textwidth]{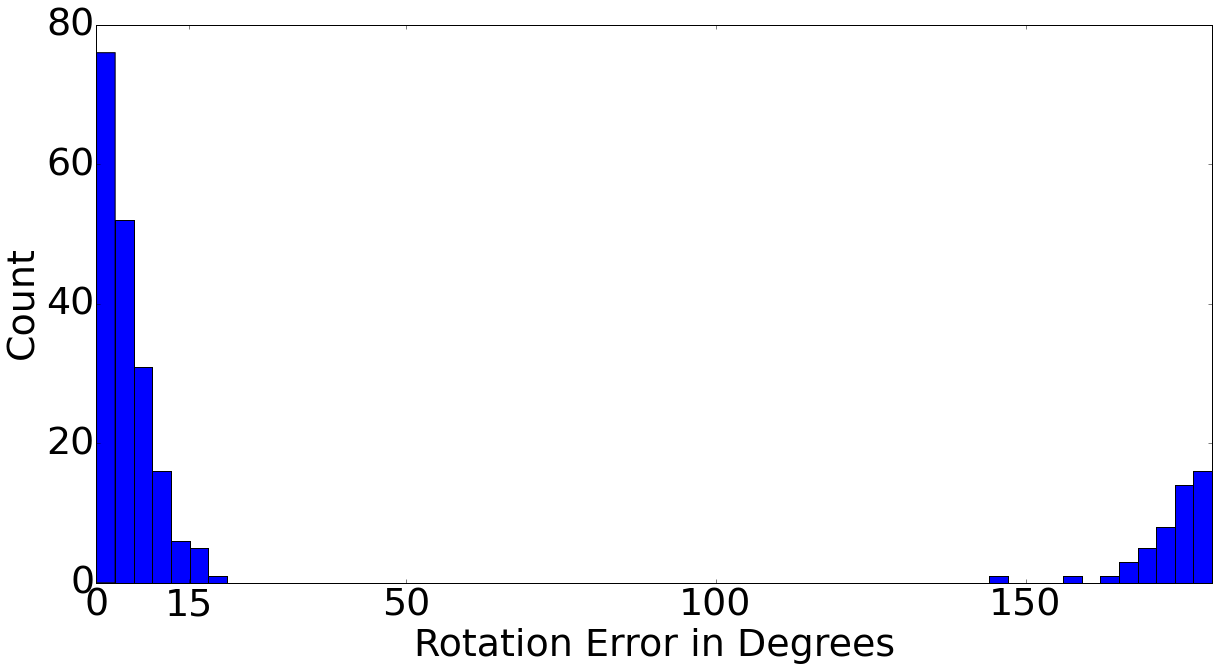}
	\caption{Histogram of rotation errors when estimating the rotation angles between $-180^\circ$ to $180^\circ$.}
	\label{google_mistake_image_right}
\end{figure}

%ignore the difference between the front and the rear of the car
We also build a baseline 3-layer CNN model following the description in \cite{henriques2016warped}, where the last layer of the network contains one node to regress the target rotation angles of the vehicles (in radians). The results are shown in Table \ref{google_mistake_image_left}. We find that the CNN model with optimal instantiations outperforms the baseline model by a big margin, and most of the rotation errors are contributed by the cases where the car fronts are mistaken for the car rears. More specifically, as we show in Figure~\ref{google_mistake_image_right}, 77\% of the data are predicted with less than $15^{\circ}$ of rotation error while 22\% are predicted with more than $150^{\circ}$ of rotation error.  If we ignore the difference between the front and the rear of the car and relax our problem by estimating the rotation angles between $-90^\circ$ to $90^\circ$, we achieve an average test rotation error of $4.87^\circ$.

Note that the CNN for regressing the rotation angle result from \cite{henriques2016warped} shown in Table~\ref{google_mistake_image_left} uses a different approach to calculate the rotation errors. Let us denote by $\alpha_i, \hat{\alpha}_i \in (-\pi, \pi)$  the ground truth and the predicted value of the angle respectively for example $i$. Henriques et al. \cite{henriques2016warped} define the rotation error as $e_i=\frac{\pi}{2}-\Big{\lvert}\lvert\alpha_i\bmod\frac{\pi}{2}-\hat{\alpha_i}\bmod\frac{\pi}{2}\rvert-\frac{\pi}{2}\Big{\rvert}$. We believe a better metric would be $e_i=\pi-\Big{\lvert}\lvert\alpha_i-\hat{\alpha_i}\rvert\bmod2\pi-\pi\Big{\rvert}$ if $\alpha_i, \hat{\alpha}_i \in (-\pi, \pi)$, and $e_i=\frac{\pi}{2}-\Big{\lvert}\lvert\alpha_i-\hat{\alpha_i}\rvert\bmod\pi-\frac{\pi}{2}\Big{\rvert}$ if $\alpha_i, \hat{\alpha}_i \in (-\pi/2, \pi/2)$. We provide the error result of the CNN for regression based on our calculation.

\begin{table}[h!]
	\centering
	\scalebox{0.95}{
		{\renewcommand{\arraystretch}{1.4}
			\begin{tabular}{|c|c|}
				\hline
				{Model Description}	 & {Average rotation error (degree)} \\ \hline
				CNN for regression \cite{henriques2016warped} & 28.87 \\
				Warped-CNN \cite{henriques2016warped} & 26.44\\\hline
				CNN for regression ($-180^\circ$ to $180^\circ$) & 63.7 \\
				CNN for regression ($-90^\circ$ to $90^\circ$) & 43.1 \\
				CNN with DC-ESGD ($-180^\circ$ to $180^\circ$) & 37.8 \\
				CNN with DC-ESGD ($-90^\circ$ to $90^\circ$) & 4.87\\ \hline
		\end{tabular}} %\caption{(a)}
	}
	\caption{The average rotation errors of different models.}
	%As we can see, most of the rotation errors are contributed by the cases in which the car fronts are mistaken for the rears.
	\label{google_mistake_image_left}
\end{table}

%In this document, we extend our previous work on deep learning with optimal instantiations and show two applications of this approach.

\section{Robustness to Clutter}\label{clutter}

\begin{figure}[H]
	\centering
	\includegraphics[width=0.8\textwidth]{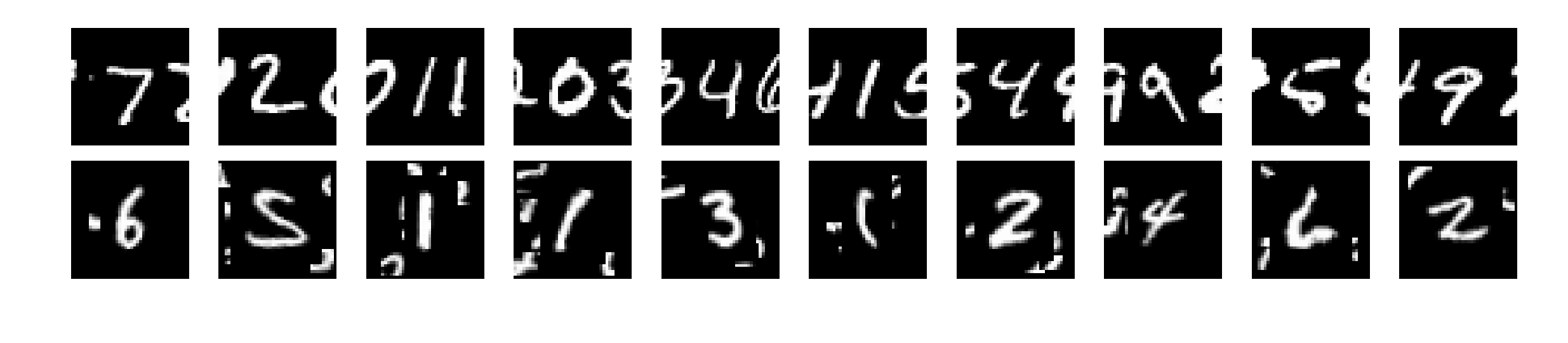}
	\caption{Sample images of two different types of clutters: flanking digits (top) and random clutter (bottom).}
	\label{digit_clutters}
\end{figure}

Handling clutter in an image is of paramount importance, as clutter can lead to significant degradation of classifier
performance if not observed during training. We thus investigate the sensitivity of the DC approach to different types of clutter, which are not observed in the training data. We employ two types of clutter models.
\begin{itemize}
	\item[]{\textbf{Flanking digits:} We put two digits on the two sides of the original digit and crop the image, so we have parts of the flanking digits as clutter. This kind of clutter is very common when dealing with digit sequence recognition.}
	\item[]{\textbf{Random clutter:} We randomly select small image patches from digit images and place them randomly around the original digits. These patches contain digit parts such as strokes and curvatures.}
\end{itemize}

\begin{figure}[b]
		\centering
		\includegraphics[width=0.8\textwidth]{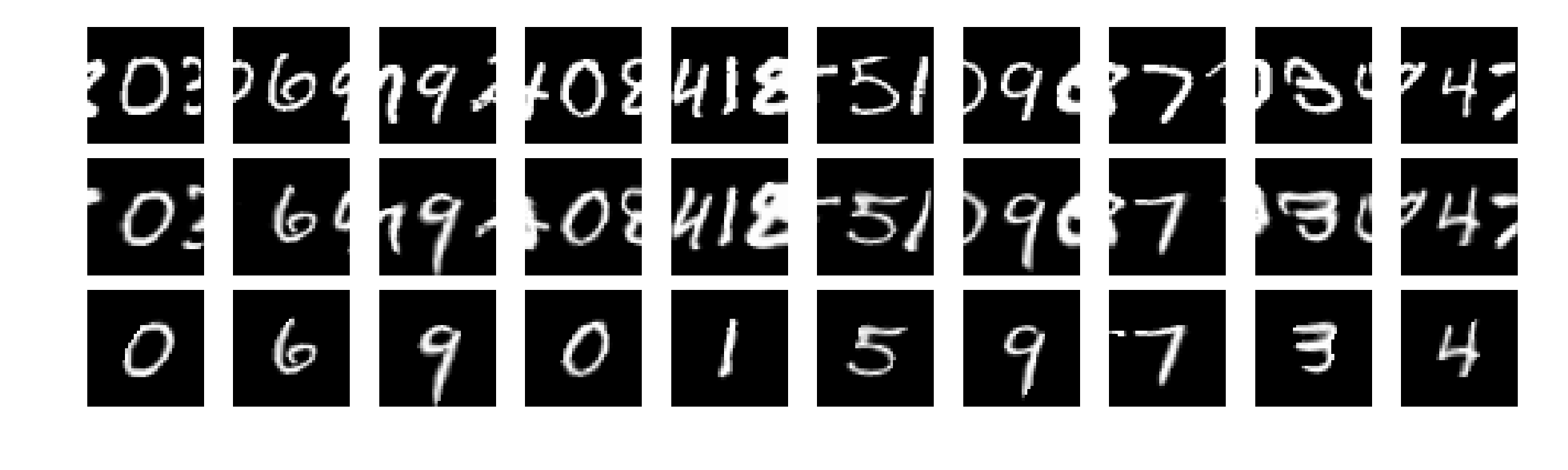}
		\caption{Examples of the original images with flanking digit clutter are shown in the first row. The corresponding recovered reference poses under the correct class labels are shown in the second row. The decluttered images extracted by applying object class support are shown in the third row using a decluttering approach described in Section \ref{decluttering_with_support_map}}.
		\label{digit_tps_clutter_flanking}
	\end{figure}

In Figure~\ref{digit_clutters} we show some examples of images with the two different clutter types. Note that nearby clutter will not touch or overlap with the original digit in the center. In the first two rows of Figure \ref{digit_tps_clutter_flanking} are the reference poses recovered by our approach for images with flanking digit clutter . Only the reference poses of the images under the correct class labels are shown here. The model is able to adjust the center digits to obtain the preferred poses. It is worth noting that, the digits in our training data have the same size as the digits in the test data. 
%Therefore, the size of the pose-adjusted test digits should remain the same, and our model is not able to remove the nearby clutter by zooming in on the center digits in the images. 
%If the model is trained on images with rescaled objects that occupy the entire canvas without any padding, then the model is able to %zoom in on the center digits and remove the surrounding clutter in the test images.

\begin{figure}[h!]
	\centering
	\includegraphics[width=0.8\textwidth]{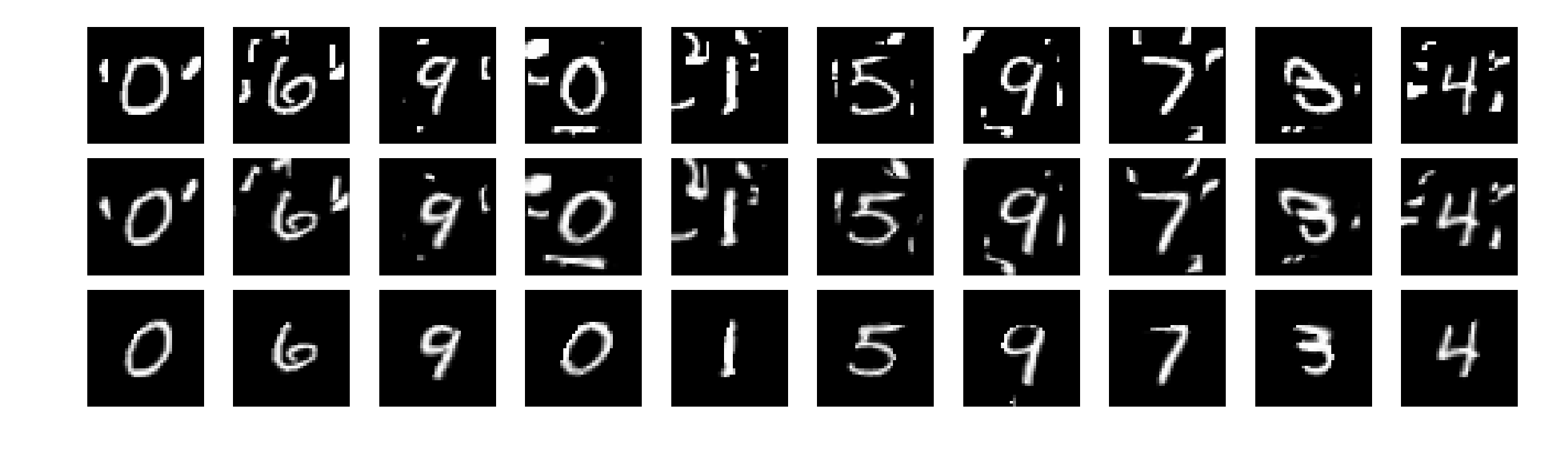}
	\caption{Examples of the original images with random clutter are shown in the first row. The corresponding recovered reference poses under the correct class labels are shown in the second row. The decluttered images extracted by applying object class support of the correct class are shown in the third row using a decluttering approach described in Section \ref{decluttering_with_support_map}.}
	\label{digit_tps_clutter_random}
\end{figure}

In the first two rows of Figure~\ref{digit_tps_clutter_random} we  show the reference poses recovered by our approach for images with random clutter surrounding the target objects. Similarly, the model can adjust the pose of the target object in the center regardless of the surrounding random clutter. It is worth noting that the reference poses estimated for the objects with surrounding random clutter are different from those captured for the objects with flanking digits. 
As we will discuss in the following section, we will show some evidence to prove that our approach is less robust to random clutter and the reference poses captured here are not perfect.

\subsection{Declutter Images with Object Class Support Map} \label{decluttering_with_support_map}
After the reference poses are estimated, the surrounding clutter still exists and can affect the out come of the
classifier. The DC framework is then used to estimate support masks for the objects, which are used to eliminate the clutter.

\begin{figure}[h]
	\centering
	\includegraphics[width=0.8\textwidth]{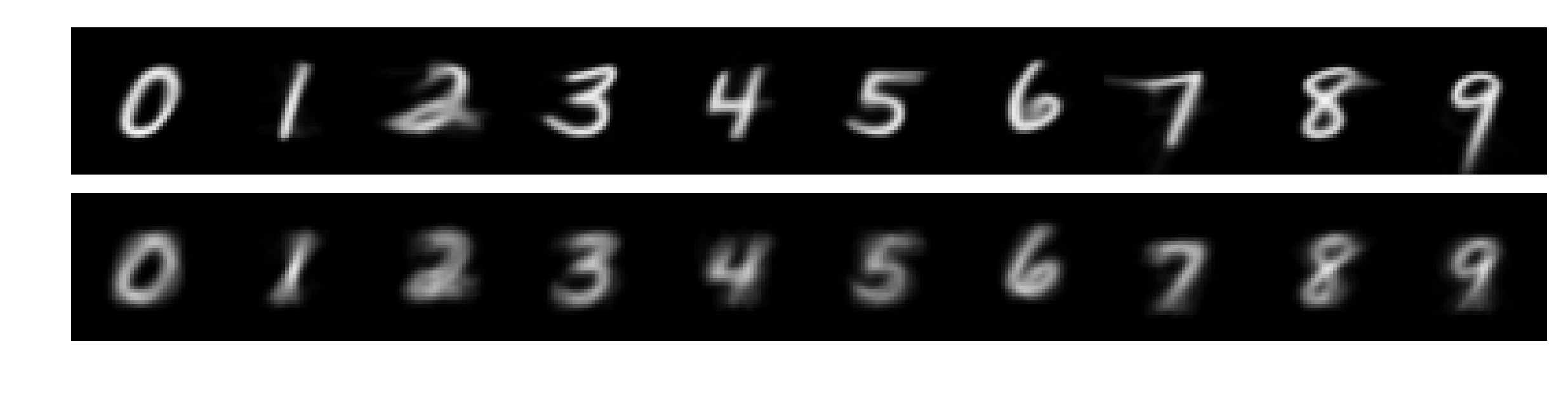}
	\caption{Mean images of handwritten digits (bottom) and pose-adjusted handwritten digits (top).}
	\label{digit_support}
\end{figure}

In Figure~\ref{digit_support}, we show the comparison between the mean images of the original handwritten digits in the training set and the pose-adjusted handwritten digits recovered by the thin plate splines (TPS). As nuisance transformations in the data are removed to obtain the reference pose of the object, it is clear that the mean images of the pose-adjusted digits are much sharper than the mean images of the original digits and can be used to determine an object support map.

In the presence of clutter, if the object labels of the images are known, we can apply the support maps of the correct class to the pose-aligned images with clutter and obtain the decluttered images, as shown in Figure~\ref{digit_tps_clutter_flanking} and Figure~\ref{digit_tps_clutter_random}. Note that this decluttering step is naturally achieved with the pre-trained model. This is very useful when dealing with tasks where the objects in the training images have clean background while the objects in the testing images are surrounded by clutter. 

Note that in Figure~\ref{digit_tps_clutter_random}, we observe that some parts of the objects are cut out by the support masks (For example, digit 9 in the seventh column and digit 7 in the eighth column). The shapes of the recovered reference poses cannot completely match the support masks of the corresponding class, indicating that the surrounding clutter is affecting the classification result.

.
%\subsection{Classify the Decluttered Images}
%We now show how decluttering images with object class support masks can be incorporated into our framework to solve classification tasks for images with clutter. Intuitively, it would be ideal if we can remove the clutter in the images before we classify them. As we showed above, we are able to obtain the support maps for each object class and use the support maps to remove the clutter in the images if the labels of the images are known. However, how can we apply the support maps to declutter a test image when the label of the target object is unknown?

\subsection{Classifying with clutter using object support maps}
The results shown in the  Figures \ref{digit_tps_clutter_flanking} and \ref{digit_tps_clutter_random} assume knowledge of the correct class. This is not known in the actual classification setting.
Since in our approach we estimate the reference pose for each class separately we can apply the support map of that class before passing it to the downstream network to get the output score for that class. The class output with highest value yields the final classification. When trained on the MNIST1000 dataset with a clear background and tested on original test dataset with flanking digit clutter, our approach improves the classification accuracy rate from $89.82\%$ to $91.07\%$ when we remove the clutter from test images using the support maps. However, if we test the model on the original test dataset with random surrounding clutter, the classification accuracy rate drops from $88.59\%$ to $86.91\%$. This again shows that our approach is less robust to random surrounding clutter.

\begin{figure}[h]
	\centering
	\includegraphics[width=\textwidth]{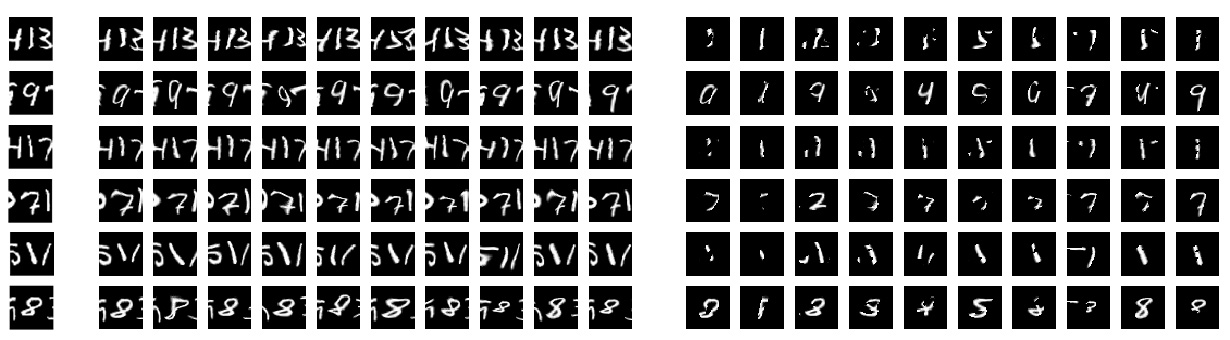}
	\caption{We show examples of misclassified digits and the corresponding optimal images captured for the corresponding images for different classes (in the middle). On the right, we show the corresponding decluttered images that we feed to the downstream network to produce class scores for classification.}
	\label{mistake_supports}
\end{figure}

The mistakes made by the decluttering approach are shown in Figure~\ref{mistake_supports}. There are two types of issues: 

\begin{itemize}
\item The subset problem:  one class, say $c$, after transformation, can look like a subset of another class $d$. 
On an image of class $d$, once the support map for $c$ is applied only the subset is visible and the image gets a high score
for class $c$.
For instance, in the second row of Figure~\ref{mistake_supports}, a digit nine can look like digit zero, digit four and digit seven after we deform it and apply the support maps. Note that our classification model will produce a class score for each class separately and label the test example with the class that has the highest score. Therefore, having decluttered images look like images from a different class other than the target class during the classification stage would confuse the classifier. 

\item Some mistakes are caused by some undesired deformations. For instance, in the examples we show in the first row of Figure~\ref{mistake_supports}, we observe in the six column that clutter from the nearby region gets pulled to the digit  in the center to form a new object that looks like a digit five. After we apply the support map to the image and remove the clutter, this image looks exactly like a digit five. This is caused by too much flexibility of the deformation allowed in the thin-plate spline, which we can alleviate by regularizing on the degree of deformation.
\end{itemize}

As explained above, for each image, we directly feed the decluttered images for different classes to the downstream classifier for classification. The class scores are produced for each decluttered image separately, and the class that produces the highest score will be picked to determine the label of the example. Since the classification model only observes the decluttered image for a certain class, without being aware of the decluttered images for other classes or what got masked out in the original image, the model simply does not have the information on whether a certain object class can best explain the scene in the original image. 

% YALI: This training is not done with clutter - I HOPE.

\begin{figure}[H]
	\centering
	\includegraphics[width=0.6\textwidth]{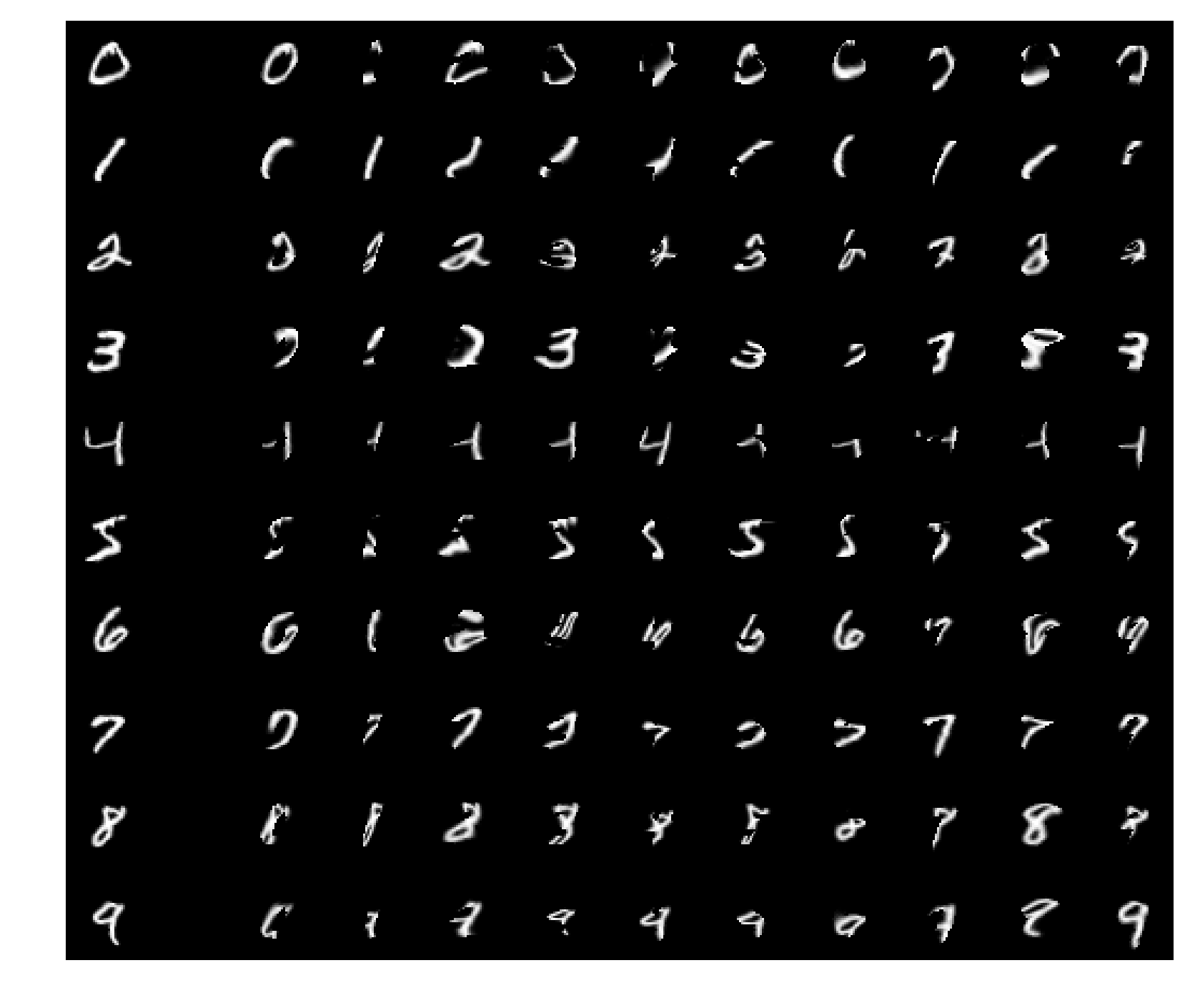}
	\caption{The stack of ten images produced from each training image, after the optimal instantiation is computed for each class and the corresponding class support mask is applied. The leftmost column is the original training image.}
	\label{train_supports}
\end{figure}

To resolve this, we apply a two-step mechanism for training and testing the images, {\it with no clutter observed in training.}
We train a regular deformable classifier $f$ and estimate a support map for each class.
Then for each training image  the optimal instantiation for each class is computed using $f$ and the corresponding support 
map applied yielding
ten transformed and cropped images, as show in Figure~\ref{train_supports}. For each training image we stack these ten transformed and cropped images and train a regular CNN to classify the label of the example based on the stack of input images. This time, since the model is able to observe the optimal deformed and decluttered images from all the classes, it has richer information on what gets masked out by the support maps, and it can better resolve the subset problem. By applying this two-step mechanism to classifying images with clutter, we achieve a classification accuracy of 93.47\% on the test images with flanking digit clutter and a classification accuracy of 91.86\% on the test images with random surrounding clutter. These are respectively 2.3\% and 4.95\% higher than the approach without the two-step mechanism.

\section{Discussion}\label{conc}

In this work, we propose a framework for training deep neural networks with optimal instantiations of the data. By introducing latent variables to parametrize the transformation of the data for each class, our approach is able to  
obtain the reference pose for the object that is being classified, and consequently can achieve better classification rates
with smaller training sets. We show that such an approach can be applied to any existing neural network architecture and is compatible with general types of transformations including rotation, translation, scaling and local deformations.
This presents a non-generative approach to estimating latent transformations, but can be used to estimate templates for 
the different classes by averaging over the pose-corrected images. When generative methods are used for classification
the templates are needed in order to compute the likelihood of each class for a given test image.
Here the
templates are not needed for classification but can be used to estimate the object support and to identify object parts. Furthermore,
by introducing discrete latent variables we believe it should be possible to  estimate clusters or mixture components for the different object classes, thus refining the estimated templates. One clear advantage of generative modeling is that for each only examples of that class
are needed to estimate the template and the distribution over the latent variables. The disadvantage of such modeling is the inadequacy of the noise models, which typically need to assume conditional independence in order for the model to be computationally tractable.
In our setting all class labels need to be known in order to update the parameters of the network. This is essentially determined by the  particular multi-class hinge loss we use. 
We note that it is also possible to use `one-against-the-rest'  hinge losses, where for each class we are estimating a two class
classifier. In that case we would be learning the deformable classifier for each class separately, and implicitly a template for that class and even a distribution over deformations,
avoiding the need to provide a generative model for the images. To summarize, much of the important information about the distribution of samples in each class that is obtained from generative modeling can be obtained in the framework proposed here, provided there is a classification cost and data available to evaluate this cost.

\newpage
\bibliographystyle{alpha}
\bibliography{bibliography.bib,book.bib}

\newcommand{\etalchar}[1]{$^{#1}$}
\begin{thebibliography}{MVKT16}

\bibitem[AAT07]{alla}
S.~Allassonni\'ere, Y~Amit, and A.~Trouv\'e.
\newblock Toward a coherent statistical framework for dense deformable template
  estimation.
\newblock {\em JRSS (Series B)}, vol. 69:3--29, 2007.

\bibitem[AGP91]{amit-jasa}
Y.~Amit, U.~Grenander, and M.~Piccioni.
\newblock Structural image restoration through deformable template.
\newblock {\em Journal of the American Statistical Association},
  86(414):376--387, 1991.

\bibitem[AT07]{amit-trouve}
Y.~Amit and A.~Trouv\'e.
\newblock Pop: Patchwork of parts models for object recognition.
\newblock {\em Intl. Jour. of Comp. Vis.}, 75:267--282, 2007.

\bibitem[ATH03]{andrews2003support}
Stuart Andrews, Ioannis Tsochantaridis, and Thomas Hofmann.
\newblock Support vector machines for multiple-instance learning.
\newblock {\em Advances in neural information processing systems}, pages
  577--584, 2003.

\bibitem[Boo89]{bookstein1989principal}
Fred~L. Bookstein.
\newblock Principal warps: Thin-plate splines and the decomposition of
  deformations.
\newblock {\em IEEE Transactions on pattern analysis and machine intelligence},
  11(6):567--585, 1989.

\bibitem[CW16a]{cohen2016group}
Taco~S Cohen and Max Welling.
\newblock Group equivariant convolutional networks.
\newblock {\em arXiv preprint arXiv:1602.07576}, 2016.

\bibitem[CW16b]{cohen2016steerable}
Taco~S Cohen and Max Welling.
\newblock Steerable cnns.
\newblock {\em arXiv preprint arXiv:1612.08498}, 2016.

\bibitem[DDFK16]{dieleman2016exploiting}
Sander Dieleman, Jeffrey De~Fauw, and Koray Kavukcuoglu.
\newblock Exploiting cyclic symmetry in convolutional neural networks.
\newblock {\em arXiv preprint arXiv:1602.02660}, 2016.

\bibitem[DSRB14]{dosovitskiy2014discriminative}
Alexey Dosovitskiy, Jost~Tobias Springenberg, Martin Riedmiller, and Thomas
  Brox.
\newblock Discriminative unsupervised feature learning with convolutional
  neural networks.
\newblock In {\em Advances in Neural Information Processing Systems}, pages
  766--774, 2014.

\bibitem[DWD15]{dieleman2015rotation}
Sander Dieleman, Kyle~W Willett, and Joni Dambre.
\newblock Rotation-invariant convolutional neural networks for galaxy
  morphology prediction.
\newblock {\em Monthly notices of the royal astronomical society},
  450(2):1441--1459, 2015.

\bibitem[FGMR10]{felzenszwalb2010object}
Pedro~F Felzenszwalb, Ross~B Girshick, David McAllester, and Deva Ramanan.
\newblock Object detection with discriminatively trained part-based models.
\newblock {\em IEEE transactions on pattern analysis and machine intelligence},
  32(9):1627--1645, 2010.

\bibitem[FGP06]{fasel2006rotation}
Beat Fasel and Daniel Gatica-Perez.
\newblock Rotation-invariant neoperceptron.
\newblock In {\em Pattern Recognition, 2006. ICPR 2006. 18th International
  Conference on}, volume~3, pages 336--339. IEEE, 2006.

\bibitem[GCK91]{grenander-chow-keenan}
U.~Grenander, Y.~Chow, and D.M. Keenan.
\newblock {\em A Pattern Theoretical Study of Biological Shape}.
\newblock Springer Verlag, New York, 1991.

\bibitem[GD14]{gens2014deep}
Robert Gens and Pedro~M Domingos.
\newblock Deep symmetry networks.
\newblock In {\em Advances in neural information processing systems}, pages
  2537--2545, 2014.

\bibitem[GM98]{grenmil}
U.~Grenander and I.~M. Miller.
\newblock Computational anatomy: an emerging discipline.
\newblock {\em Quarterly of Applied Mathematics}, LVI(4):617--694, 1998.

\bibitem[Gre93]{grenander-book}
U.~Grenander.
\newblock {\em General Pattern Theory}.
\newblock Oxford University Press, Oxford, 1993.

\bibitem[HK08]{heitz2008learning}
Geremy Heitz and Daphne Koller.
\newblock Learning spatial context: Using stuff to find things.
\newblock In {\em European conference on computer vision}, pages 30--43.
  Springer, 2008.

\bibitem[HMCB14]{henriques2014fast}
Jo{\~a}o~F Henriques, Pedro Martins, Rui~F Caseiro, and Jorge Batista.
\newblock Fast training of pose detectors in the fourier domain.
\newblock In {\em Advances in neural information processing systems}, pages
  3050--3058, 2014.

\bibitem[HV16]{henriques2016warped}
Jo{\~a}o~F Henriques and Andrea Vedaldi.
\newblock Warped convolutions: Efficient invariance to spatial transformations.
\newblock {\em arXiv preprint arXiv:1609.04382}, 2016.

\bibitem[JSZ{\etalchar{+}}15]{jaderberg2015spatial}
Max Jaderberg, Karen Simonyan, Andrew Zisserman, et~al.
\newblock Spatial transformer networks.
\newblock In {\em Advances in Neural Information Processing Systems}, pages
  2017--2025, 2015.

\bibitem[KH09]{krizhevsky2009learning}
Alex Krizhevsky and Geoffrey Hinton.
\newblock Learning multiple layers of features from tiny images.
\newblock 2009.

\bibitem[LCB98]{lecun1998mnist}
Yann LeCun, Corinna Cortes, and Christopher~JC Burges.
\newblock The mnist database of handwritten digits, 1998.

\bibitem[LEC{\etalchar{+}}07]{larochelle2007empirical}
Hugo Larochelle, Dumitru Erhan, Aaron Courville, James Bergstra, and Yoshua
  Bengio.
\newblock An empirical evaluation of deep architectures on problems with many
  factors of variation.
\newblock In {\em Proceedings of the 24th international conference on Machine
  learning}, pages 473--480. ACM, 2007.

\bibitem[LSBP16]{laptev2016ti}
Dmitry Laptev, Nikolay Savinov, Joachim~M Buhmann, and Marc Pollefeys.
\newblock Ti-pooling: transformation-invariant pooling for feature learning in
  convolutional neural networks.
\newblock In {\em Proceedings of the IEEE Conference on Computer Vision and
  Pattern Recognition}, pages 289--297, 2016.

\bibitem[MTY06]{MTY}
M.~I. Miller, A.~Trouv\'e, and L.~Younes.
\newblock Geodesic shooting for computational anatomy.
\newblock {\em J. Math Imaging Vis.}, page 209Ð228, 2006.

\bibitem[MVKT16]{marcos2016rotation}
Diego Marcos, Michele Volpi, Nikos Komodakis, and Devis Tuia.
\newblock Rotation equivariant vector field networks.
\newblock {\em arXiv preprint arXiv:1612.09346}, 2016.

\bibitem[RMC15]{radford2015unsupervised}
Alec Radford, Luke Metz, and Soumith Chintala.
\newblock Unsupervised representation learning with deep convolutional
  generative adversarial networks.
\newblock {\em arXiv preprint arXiv:1511.06434}, 2015.

\bibitem[SL12]{sohn2012learning}
Kihyuk Sohn and Honglak Lee.
\newblock Learning invariant representations with local transformations.
\newblock {\em arXiv preprint arXiv:1206.6418}, 2012.

\bibitem[TH16]{teney2016learning}
Damien Teney and Martial Hebert.
\newblock Learning to extract motion from videos in convolutional neural
  networks.
\newblock {\em arXiv preprint arXiv:1601.07532}, 2016.

\bibitem[vNP17]{van2017learning}
Nanne van Noord and Eric Postma.
\newblock Learning scale-variant and scale-invariant features for deep image
  classification.
\newblock {\em Pattern Recognition}, 61:583--592, 2017.

\bibitem[WHK15]{wu2015flip}
Fa~Wu, Peijun Hu, and Dexing Kong.
\newblock Flip-rotate-pooling convolution and split dropout on convolution
  neural networks for image classification.
\newblock {\em arXiv preprint arXiv:1507.08754}, 2015.

\bibitem[You10]{younes}
L.~Younes.
\newblock {\em Shapes and Deformations}.
\newblock Springer, 2010.

\bibitem[ZYQJ17]{zhou2017oriented}
Yanzhao Zhou, Qixiang Ye, Qiang Qiu, and Jianbin Jiao.
\newblock Oriented response networks.
\newblock {\em arXiv preprint arXiv:1701.01833}, 2017.

\end{thebibliography}
\end{document}